\documentclass[conference]{IEEEtran}
\IEEEoverridecommandlockouts

\usepackage{cite}
\usepackage{amsmath,amssymb,amsfonts}
\usepackage{algorithmic}
\usepackage{graphicx}
\usepackage{textcomp}
\usepackage{xcolor}
\usepackage{url}
\usepackage{booktabs}
\usepackage{multirow}

\def\BibTeX{{\rm B\kern-.05em{\sc i\kern-.025em b}\kern-.08em
    T\kern-.1667em\lower.7ex\hbox{E}\kern-.125emX}}

\begin{document}

\title{Data-Efficient Algal Bloom Segmentation in Freshwater Bodies Using a
Three-Stage Transfer Learning Pipeline with Attention-Enhanced Decoding}

\author{
\IEEEauthorblockN{Albert Ansah Benson, Mingqiang Han, Rahul Harsha Cheppally,\\
Treman Singh Kaloya, Menard Soni, Sudan Baral}
\IEEEauthorblockA{Department of Biological and Agricultural
Engineering, Kansas State University\\
\{aabenson, mingqiang, r4hul, tremankaloya, masoni, sudanb\}@ksu.edu}
}

\maketitle

\begin{abstract}
Harmful algal blooms threaten water quality and public health, yet automated segmentation of bloom regions in freshwater imagery is limited by scarce pixel-level annotations. We present a transfer-learning pipeline that adapts a Swin Transformer U-Net with SCSE decoder attention from ImageNet-1K to water-body segmentation (RIWA, LaRS) and then to algal blooms, together with a model-assisted annotation protocol that expands the labelled algae corpus from 145 to 655 images. In a single-seed ablation on a 41-image validation split (checkpoint S2-A4 and its variants), corpus expansion is the dominant factor ($+$11 mIoU points); water-body pretraining helps only on the expanded corpus ($+$1.1 points over ImageNet initialization); SCSE attention adds 0.3 points; and a learnable positional encoding lowers mIoU by 0.6 points and is omitted. Adding 476 images, 188 of them algae-free, reduces false-positive area on non-water scenes from 22.6\% to 4.9\%. Against nine baselines on a 256-image test split, a separately trained checkpoint of the same architecture (S2-A13) reaches mIoU 0.919 and the highest algae IoU (0.902); a near-duplicate audit indicates train/test overlap in this split, so these scores are likely optimistic.
\end{abstract}


%% ============================================================
\section{Introduction}
%% ============================================================

Freshwater ecosystems are under mounting anthropogenic pressure from
agricultural runoff, urban wastewater discharge, and rising water temperatures
associated with climate change~\cite{nwankwegu_harmful_2019,smucker_role_2023}.
A particularly visible and damaging consequence is the proliferation of algal
blooms --- the rapid, often explosive growth of algae or cyanobacteria on the
surface of lakes, reservoirs, rivers, and other inland water
bodies~\cite{shi_anthropogenic_2026}. Surface algal blooms degrade water quality
through the production of toxins harmful to humans and aquatic fauna, the
depletion of dissolved oxygen as bloom biomass decomposes, and the physical
obstruction of light penetration that disrupts subaquatic food web
dynamics~\cite{qin_review_2025}. The consequences extend beyond ecology:
impaired water sources affect drinking water supply, irrigation, fisheries, and
recreational use~\cite{dodds2009eutrophication,noaa_hitting_2022}, placing algal
bloom monitoring at the intersection of environmental science and public health
policy.

Effective management of algal bloom events demands timely, spatially resolved
information about bloom extent, density, and progression. Traditional monitoring
approaches rely on field sampling, laboratory analysis, and satellite remote
sensing. Satellite imagery is constrained by coarse spatial resolution, cloud
cover interference, and temporal resolutions that limit its utility for
early-warning applications in dynamic bloom
events~\cite{shen2012satellite,stumpf_satellite_2021,liu_harmonized_2025}.
Field sampling provides ground truth but is labour-intensive, geographically
sparse, and not scalable to the temporal demands of continuous monitoring.
Drone-based (UAV) imaging has emerged as a compelling middle ground: capable of
delivering high-resolution, on-demand imagery at low operational cost, UAVs can
survey water body surfaces with sufficient spatial detail to capture early-stage
bloom patches that satellite sensors would fail to
resolve~\cite{wu_review_2019,kislik2018uavs,shang_remote_2020}.

Semantic segmentation --- the task of assigning a class label to every pixel ---
provides a natural computational framework for quantifying algal surface coverage
from drone imagery. Deep learning-based segmentation models, particularly those
built on encoder-decoder architectures with large-capacity pretrained backbones,
have achieved remarkable performance across remote sensing and biological imaging
domains~\cite{huang_deep_2023}. However, their application to algal bloom
detection confronts a fundamental bottleneck: the severe scarcity of pixel-level
annotated training data~\cite{wang_deep_2023}. Unlike established agricultural
computer vision benchmarks where thousands of annotated images are publicly
available, no large-scale public datasets provide dense segmentation masks for
algal bloom imagery. Manually annotating such data is prohibitively expensive,
requiring domain expertise to reliably distinguish algae from visually similar
background classes --- open water, sky reflections, submerged aquatic vegetation,
and riparian growth --- that vary substantially across imaging conditions, seasons,
and geographic locations~\cite{russwurm_semantic_2023}.

This scarcity motivates the central methodological contribution of the present
work: a three-stage transfer learning pipeline that progressively adapts a
pretrained segmentation model from general visual representations toward
water-body-specific and ultimately algae-specific features, without requiring
large quantities of directly annotated algal bloom imagery. The approach draws
conceptual inspiration from Ghosh et al.~\cite{ghosh_decoder-aware_nodate},
who demonstrated strong segmentation performance using only 99 labeled samples
in the wheat organ domain by combining self-supervised continual pretraining with
similarity-aware candidate selection and uncertainty-guided pseudo-labeling.

The proposed pipeline initializes a Swin Transformer~\cite{liu2021swin} encoder from ImageNet-1K~\cite{russakovsky2015imagenet}
pretrained weights and fine-tunes it sequentially on two water body segmentation
datasets before final adaptation to the algal bloom task. A model-assisted
annotation expansion protocol was developed in which the best-performing Stage~2
model generates initial segmentation masks for previously unannotated images,
subsequently reviewed and verified by a human annotator. A controlled ablation
study across ten experiments was then conducted to isolate the contribution
of each pipeline component, including the two architectural enhancements.

%% ============================================================
\section{Related Work}
%% ============================================================

\subsection{Semantic Segmentation for Environmental Monitoring}

Deep learning-based semantic segmentation has been increasingly applied to remote
sensing imagery for environmental monitoring tasks including land cover
classification, vegetation mapping, and water body delineation. Encoder-decoder
architectures, particularly those built on transformer backbones, have
demonstrated strong performance across diverse remote sensing domains. Multi-task
learning frameworks have further extended segmentation capabilities; WeedSense
~\cite{sarker_weedsense_2025} jointly performs semantic segmentation, height
estimation, and growth stage classification for agricultural weed analysis,
demonstrating the utility of dense prediction approaches in environmental imaging
contexts.

\subsection{Transfer Learning and Domain Adaptation in Low-Label Regimes}

Transfer learning from large-scale pretraining datasets has become the dominant
strategy for addressing annotation scarcity in specialized imaging domains.
Self-adapting vision-language models to edge
devices~\cite{leonardis_self-adapting_2025} demonstrated that large pretrained
representations can be compressed and adapted for resource-constrained
deployment. In agricultural vision, Ghosh et al.~\cite{ghosh_decoder-aware_nodate}
proposed a decoder-aware self-supervised continual pretraining framework with
uncertainty-guided pseudo-labeling achieving mIoU of 73.61 on the validation set
using only 99 labeled wheat organ samples, directly motivating the data-efficient
transfer approach adopted in this work. The extent to which domain-specific
pretraining transfers across closely related but semantically distinct tasks ---
such as water body segmentation and algal bloom segmentation --- remains an open
empirical question that this work directly addresses.

\subsection{Attention Mechanisms in Segmentation Decoders}

Channel and spatial attention mechanisms have been widely adopted to improve
feature selectivity in segmentation architectures. Squeeze-and-Excitation (SE)
networks~\cite{hu_squeeze-and-excitation_2019} introduced channel-wise feature
recalibration through global average pooling and gated bottleneck projections.
Concurrent spatial and channel squeeze-and-excitation (SCSE)~\cite{roy2018concurrent} extends this to
jointly recalibrate channel and spatial responses, improving sensitivity to
small or spatially sparse target regions --- a property directly relevant to
the detection of early-stage algal bloom patches. In the present work, SCSE
attention is incorporated into the decoder on this basis;
Section~\ref{sec:ablation_discussion} isolates its contribution via ablation
and finds a small benefit (+0.26 mIoU points on the expanded dataset, single
seed).

%% ============================================================
\section{Methodology}
%% ============================================================

\subsection{Overview}

This work presents a multi-stage, transfer-learning-based semantic segmentation
framework designed to detect and delineate algal bloom regions in water body
imagery. Training proceeds in three stages: from general visual features
(ImageNet-1K), to water-body-specific features (RIWA and LaRS, Stage~1), to
algal bloom segmentation (Stages~2 and~3). Stages~2 and~3 both initialize
from the Stage~1 checkpoint; the Stage~2 model additionally generates the mask
proposals used to build the larger Stage~3 training corpus.

\begin{figure*}[htbp]
\centerline{\includegraphics[width=\textwidth]{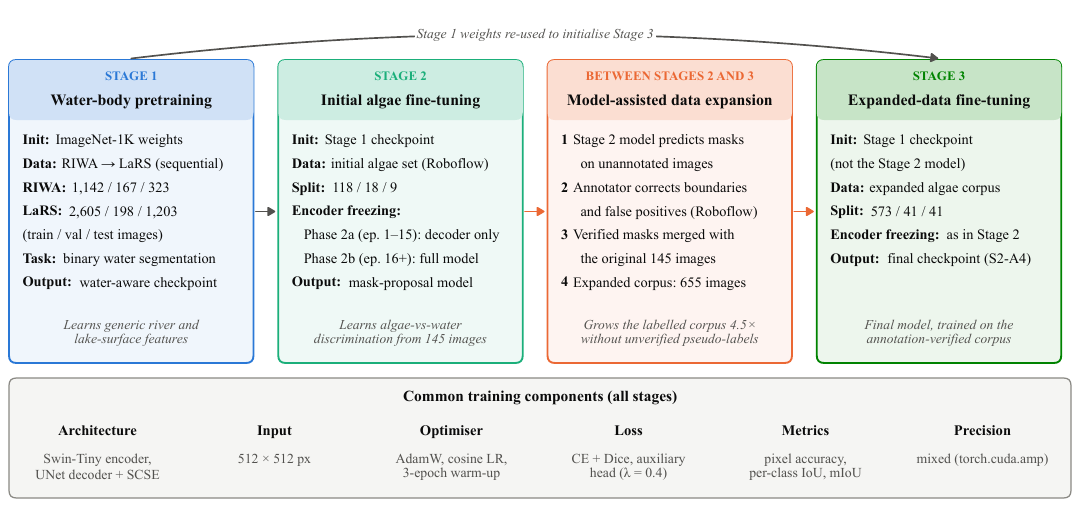}}
\caption{Overview of the three-stage transfer learning pipeline. Stage~1
pretrains on water-body segmentation (RIWA, LaRS); Stage~2 fine-tunes on the
initial 145 algae images; Stage~3 retrains from the Stage~1 weights on the
655-image corpus produced by model-assisted annotation (orange).}
\label{fig:pipeline}
\end{figure*}

\begin{figure*}[htbp]
\centerline{\includegraphics[width=\textwidth]{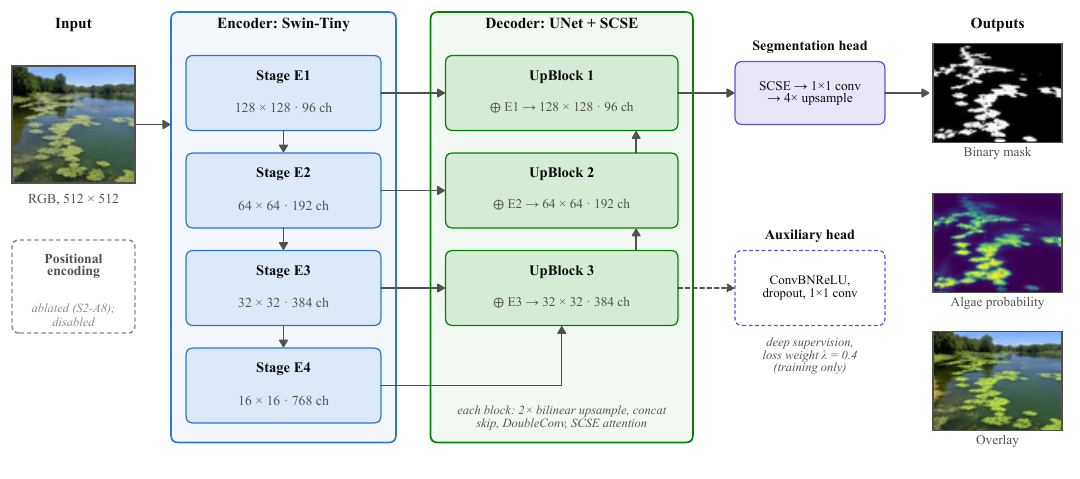}}
\caption{Attention-enhanced Swin-UNet. Skip connections link encoder stages
E1--E3 to SCSE-equipped UpBlocks; the auxiliary head is used only in training.
The positional encoding (dashed) is disabled (Section~\ref{sec:ablation}).}
\label{fig:architecture}
\end{figure*}

Two architectural enhancements accompany the training curriculum: (i) SCSE
attention~\cite{roy2018concurrent} integrated into every decoder
upsampling block to encourage sensitivity to small and spatially sparse bloom
patches; and (ii) a learnable spatial positional encoding module prepended to
the encoder input to supply explicit coordinate context. Section~\ref{sec:ablation_discussion}
isolates the contribution of each enhancement via ablation (S2-A5, S2-A8). A model-assisted dataset expansion
protocol (Section~\ref{sec:expansion}) addresses the annotation scarcity bottleneck.
Fig.~\ref{fig:pipeline} provides a comprehensive overview and
Fig.~\ref{fig:architecture} details the model architecture.

\subsection{Datasets}
\label{sec:datasets}

Table~\ref{tab:datasets} lists the data used in this work. Stage~1 uses two
public water-segmentation datasets: \textit{RIWA}~\cite{wagner2023riwa,wagner_riwa_kaggle}, river and waterway scenes
(1,632 images), and \textit{LaRS}~\cite{zust2023lars}, lake, river and sea scenes (4,006
images). The algae experiments use two separate image pools.

\textit{Ablation pool.} The initial algae dataset contains 145
Roboflow-annotated images (118/18/9 train/validation/test). Model-assisted
annotation (Section~\ref{sec:expansion}) expanded it to 655 images
(573/41/41); every ablation experiment (Section~\ref{sec:ablation}) is
trained on this corpus and evaluated on its 41-image validation split. One
further checkpoint, \textbf{S2-A4$^{+}$}, was trained on a 1,049-image corpus
that adds 476 images from the validation split of
\textit{Algae-data-reshuffled} (188 of them algae-free, including non-water
scenes) to the 573 training images, in order to reduce false positives on
non-water imagery (Section~\ref{sec:model_comparison}); its validation and
test splits are unchanged.

\textit{Comparison pool.} The eight baselines and checkpoint S2-A13 are
trained on \texttt{phase2\_dataset14} (1,714 training / 487 validation
images), and all ten compared models are evaluated on the 256-image test
split of Algae-data-reshuffled, which none of them uses for training. Split
separation was checked by file identity only, which does not exclude
augmented or re-encoded copies of the same photograph;
Section~\ref{sec:model_comparison} reports a subsequent near-duplicate
audit.

\begin{table}[htbp]
\caption{Datasets and their roles.}
\label{tab:datasets}
\begin{center}
\footnotesize
\begin{tabular}{p{2.6cm}cc p{2.0cm}}
\hline
\textbf{Dataset} & \textbf{Train} & \textbf{Val/Test} & \textbf{Used for} \\
\hline
\multicolumn{4}{l}{\textit{Stage~1 --- water-body pretraining}} \\
RIWA                       & 1,142 & 167 / 323   & Stage 1 \\
LaRS                       & 2,605 & 198 / 1,203 & Stage 1 \\
\hline
\multicolumn{4}{l}{\textit{Ablation pool}} \\
Algae, initial             & 118   & 18 / 9      & S2-A1 -- S2-A3 \\
Algae, expanded            & 573   & 41 / 41     & S2-A4, A4$'$, A5, A8 \\
Algae, expanded + extra$^\dagger$ & 1,049 & 41 / 41 & S2-A4$^{+}$ \\
\hline
\multicolumn{4}{l}{\textit{Comparison pool}} \\
\texttt{phase2\_dataset14} & 1,714 & 487 / ---   & Baselines, S2-A13 \\
Algae-data-reshuffled      & ---   & --- / 256   & Comparison test \\
\hline
\end{tabular}
\end{center}
\footnotesize{$^\dagger$573 expanded training images plus 476 images from the
Algae-data-reshuffled validation split.}
\end{table}

\subsection{Model Architecture}
\label{sec:architecture}

\textit{Encoder:} A Swin Transformer~\cite{liu2021swin} (\textit{swin\_tiny\_patch4\_window7\_224})
instantiated via \texttt{timm}~\cite{wightman2019timm} with \texttt{features\_only=True} produces
hierarchical feature maps at four levels with channel dimensions
$[96, 192, 384, 768]$. All models trained at $512 \times 512$ pixels.

\textit{Decoder:} A UNet-style~\cite{ronneberger2015unet} decoder with three UpBlocks. Each UpBlock
performs bilinear upsampling, skip-connection concatenation, and DoubleConv.
Decoder channel widths follow $[384, 192, 96]$ (\textit{fpn\_dim} = 96).

\textit{SCSE Attention Module:} Incorporated into each UpBlock following
DoubleConv. Channel Squeeze-and-Excitation (cSE) and Spatial
Squeeze-and-Excitation (sSE) applied in parallel:

\begin{equation}
\mathbf{Y} = \mathbf{X} \odot \sigma(\text{cSE}(\mathbf{X})) +
             \mathbf{X} \odot \sigma(\text{sSE}(\mathbf{X}))
\label{eq:scse}
\end{equation}

An additional SCSE block is applied to the final decoder feature map before the
primary segmentation head.

\textit{Learnable Spatial Positional Encoding (SPE):} A 2-channel coordinate
grid normalized to $[-1,1]$ is projected to 3 channels via two $1{\times}1$
convolutions ($2 \to 16 \to 3$) and added element-wise to the RGB input, scaled
by a learnable scalar initialized to 0.1. An ablation study
(Section~\ref{sec:ablation_discussion}) found this module mildly
counterproductive at the current dataset scale, so it is omitted from the
final configuration (S2-A8 and S2-A4$^{+}$). Checkpoint S2-A13, used in the
model comparison (Table~\ref{tab:model_comparison}), was trained before this
ablation and still includes it.

\subsection{Three-Stage Training Pipeline}

\textit{Stage 1 --- Water Body Pretraining:} The full model was trained on RIWA
first, then continued on LaRS from the best RIWA checkpoint, adapting encoder and
decoder to aquatic imagery statistics prior to any algae-specific data.

\textit{Stage 2 --- Algae Fine-Tuning with Encoder Freezing:} A two-phase
protocol mitigates catastrophic forgetting. In Phase~2a (Epochs 1--15), the
encoder is frozen and only decoder weights, attention modules, and segmentation
heads are updated. In Phase~2b (Epochs 16 onward), all parameters are trained jointly.

\textit{Stage 3 --- Fine-Tuning on the Expanded Dataset:} Stage~2 training is
repeated from the Stage~1 checkpoint using the expanded 655-image corpus with
the same two-phase freezing protocol.

\subsection{Loss Function}

A weighted combination of pixel-wise cross-entropy (CE) and soft Dice loss~\cite{milletari2016vnet}
applied to both primary and auxiliary outputs:

\begin{equation}
\mathcal{L}_{\text{total}} = \mathcal{L}_{\text{main}} +
0.4 \cdot \mathcal{L}_{\text{aux}}
\end{equation}
\begin{equation}
\mathcal{L}_{\text{main}} = \mathcal{L}_{\text{CE}}(\hat{y}, y) +
\mathcal{L}_{\text{Dice}}(\hat{y}, y)
\end{equation}

Class weights $[1.0, 4.0]$ (background, algae) compensate for severe spatial
class imbalance. 

\subsection{Model-Assisted Dataset Expansion}
\label{sec:expansion}

The initial 145-instance dataset produced substantial overfitting during Stage~2
fine-tuning. A model-assisted labeling strategy expanded the corpus while
maintaining quality through a human in the loop: (1)~the best Stage~2 model
applied to unannotated bloom images to generate initial binary masks;
(2)~a human annotator corrected boundaries, filled missed regions, and removed
false positives using the Roboflow polygon tool; (3)~verified annotations merged
with original labeled instances to produce the 655-image expanded corpus.

\subsection{Data Preprocessing and Augmentation}
\label{sec:preprocessing}

Images were resized to $512 \times 512$ with aspect-ratio-preserving bilinear
interpolation and zero-padding; pixel values normalized to $[0, 1]$. Training
augmentations: algae-guided crop sampling (probability 0.70), random horizontal
and vertical flips, color jitter, random rotation and affine transforms, and
Gaussian blur and noise injection. A weighted random sampler oversamples images
containing algae annotations. These steps reduce sensitivity to bloom patch orientation and variation in imaging conditions.

\subsection{Implementation Details}

PyTorch with \texttt{timm} backbone instantiation. AdamW optimizer~\cite{loshchilov2019adamw}, learning
rate $5 \times 10^{-5}$, weight decay $1 \times 10^{-4}$, cosine annealing
schedule with 3-epoch linear warm-up, mixed-precision training
(\texttt{torch.cuda.amp}). Experiment tracking via Weights \& Biases.
Hyperparameters summarized in Table~\ref{tab:hyperparams}.

\begin{table}[htbp]
\caption{Key hyperparameters for Stage~2 and Stage~3 fine-tuning.}
\label{tab:hyperparams}
\begin{center}
\begin{tabular}{ll}
\hline
\textbf{Hyperparameter} & \textbf{Value} \\
\hline
Optimizer               & AdamW \\
Base learning rate      & $5 \times 10^{-5}$ \\
Weight decay            & $1 \times 10^{-4}$ \\
LR schedule             & Cosine annealing + 3-epoch warm-up \\
Training epochs         & 50 \\
Encoder freeze duration & 15 epochs (Phase~2a) \\
Batch size              & 4 \\
Input resolution        & $512 \times 512$ px \\
Backbone                & \textit{swin\_tiny\_patch4\_window7\_224} \\
Decoder channels        & $384 \to 192 \to 96$ \\
Dropout rate            & 0.1 \\
$\lambda_{\text{aux}}$  & 0.4 \\
Class weights           & $[1.0,\ 4.0]$ (bg, algae) \\
Mixed precision         & Enabled (\texttt{torch.cuda.amp}) \\
Algae crop sampling $p$ & 0.70 \\
\hline
\end{tabular}
\end{center}
\end{table}

%% ============================================================
\section{Results and Discussion}
\label{sec:results}
%% ============================================================

\subsection{Stage~1 Water Body Segmentation}

RIWA pretraining achieves mIoU~0.9295 on river and waterway scenes. Continuing
on LaRS from the RIWA checkpoint improves this to mIoU~0.9899, with both classes
exceeding IoU~0.988 and validation loss falling from 0.2432 to 0.0310 (full
per-class breakdown in Table~\ref{tab:ablation_stage1}, reported alongside the
Stage~1 ablation in Section~\ref{sec:ablation}). This near-ceiling performance
confirms that the encoder acquires highly generalised aquatic representations
across diverse scene types before any algae-specific data is introduced. The
critical question addressed in Section~\ref{sec:ablation} is whether these
representations transfer usefully to the algae segmentation task.

\subsection{Test Set Qualitative Results}
\label{sec:qualitative}

\begin{figure}[htbp]
\centerline{\includegraphics[width=0.82\columnwidth]{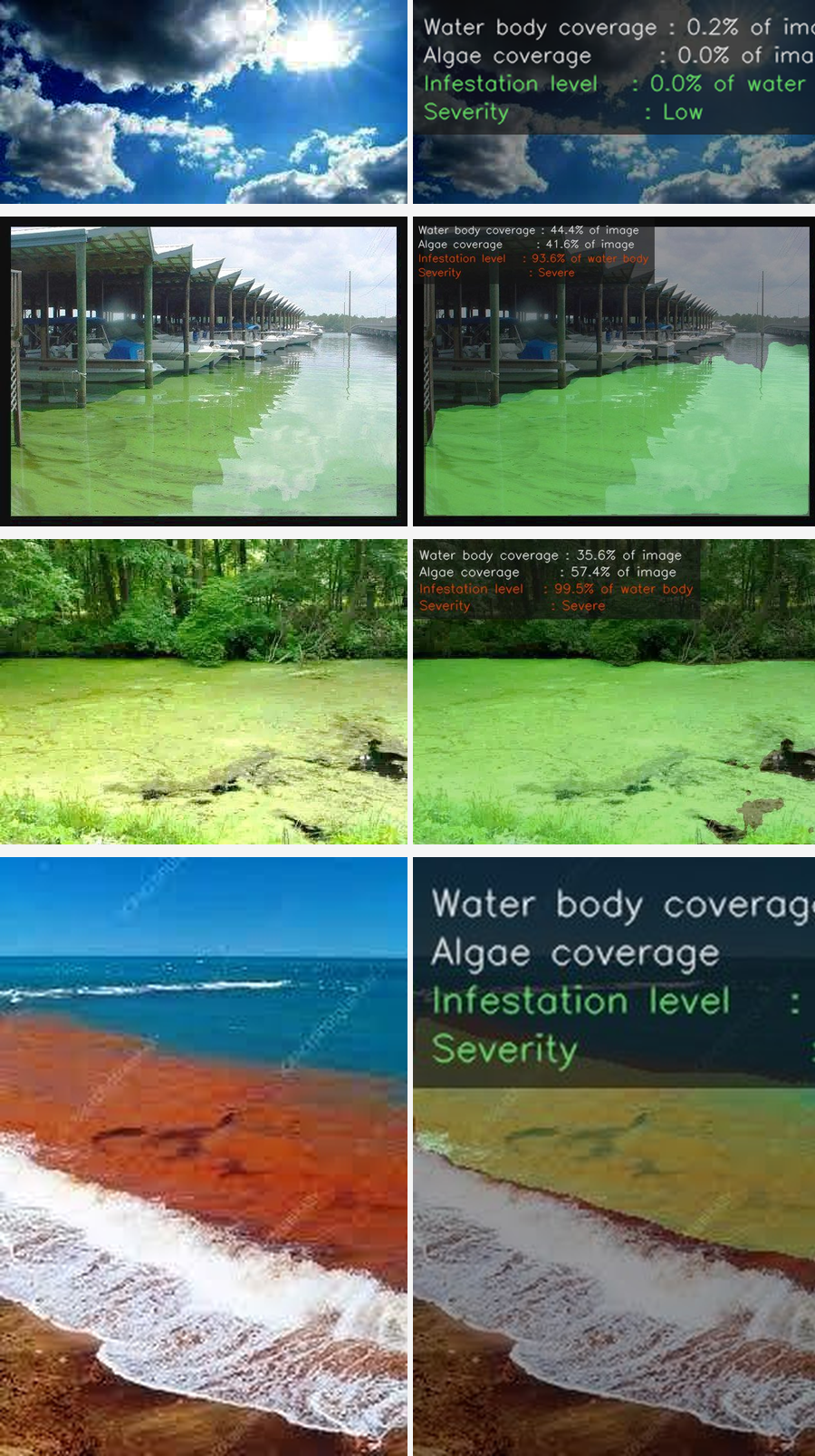}}
\caption{Qualitative results on held-out images (input, prediction): (1)~clear
sky, (2)~marina bloom, (3)~forest-pond bloom, (4)~marine red tide
(out of distribution).}
\label{fig:qualitative_test}
\end{figure}

Fig.~\ref{fig:qualitative_test} presents inference results on four held-out test
images spanning ecologically relevant scenarios. \textbf{Row~1} (clear sky): the
model correctly assigns near-zero water body coverage and Low severity,
demonstrating robust abstention from false positives in the absence of any water
surface. \textbf{Row~2} (marina cyanobacterial bloom): 93.6\% infestation
classified Severe, with the overlay correctly excluding dock structures and boat
hulls. \textbf{Row~3} (forest pond dense bloom): 99.5\% infestation classified
Severe, with background labels correctly assigned to surrounding riparian
vegetation. \textbf{Row~4} (coastal red tide): a spectrally out-of-distribution
bloom type from a marine environment, included specifically to probe
generalisation beyond the freshwater green-algae training distribution;
the model is not expected to perform optimally on such out-of-scope inputs.

%% ============================================================
\subsection{Comparison With Baseline Segmentation Models}
\label{sec:model_comparison}
%% ============================================================

Ten models were compared on the 256-image test split of
Algae-data-reshuffled: eight baselines --- U-Net~\cite{ronneberger2015unet},
UNet++~\cite{zhou2018unetpp}, FPN~\cite{lin2017fpn}, MA-Net~\cite{fan2020manet},
DeepLabV3+~\cite{chen2018deeplabv3plus} (Segmentation Models
PyTorch~\cite{iakubovskii2019smp}) and DeepLabV3~\cite{chen2017deeplabv3}, with
ResNet~\cite{he2016resnet}, EfficientNet~\cite{tan2019efficientnet} and
MobileNetV3~\cite{howard2019mobilenetv3} backbones ---
the proposed architecture (checkpoint S2-A13), and
EoMT-S~\cite{kerssies2025eomt}. The eight baselines and S2-A13 are trained on
the same \texttt{phase2\_dataset14} split. Because this comparison uses a
different data pool from the ablation study (Section~\ref{sec:datasets}), its
numbers should not be compared with Table~\ref{tab:ablation_stage2}.

\begin{figure}[htbp]
\centerline{\includegraphics[width=\columnwidth]{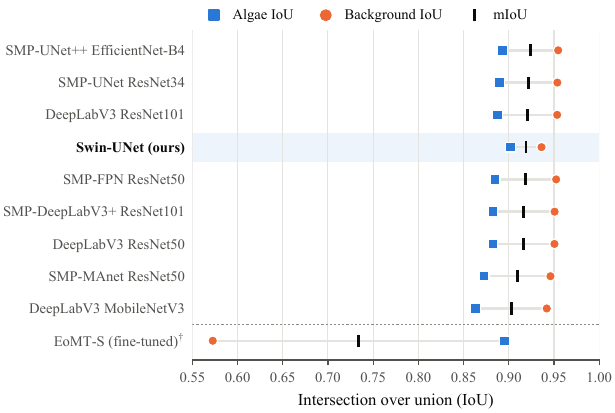}}
\caption{Per-class IoU on the 256-image comparison test set (order as in
Table~\ref{tab:model_comparison}). $^{\dagger}$Smaller training budget;
excludes background-only images.}
\label{fig:model_comparison_per_class}
\end{figure}

Table~\ref{tab:model_comparison} summarizes the results. The proposed model
ranks fourth by mIoU (0.9192), behind SMP-UNet++ (EfficientNet-B4), SMP-UNet
(ResNet34) and DeepLabV3 (ResNet101), but has the highest algae IoU (0.9021)
and algae precision (0.9499) of the ten models
(Fig.~\ref{fig:model_comparison_per_class}). The mIoU spread among the nine
fully trained models is small (0.903--0.924), and given the caveats below the
ranking among them should be regarded as tentative.

\textbf{Caveats.} Train/test separation for this comparison was verified by
file identity only. A subsequent perceptual near-duplicate audit found that
186 of the 256 test images (73\%) are exact or near-duplicates (e.g.,
augmented copies) of images in the \texttt{phase2\_dataset14} training pool,
so absolute scores are likely optimistic for all models. In addition, the
baselines received ImageNet-normalized inputs, whereas S2-A13 received inputs
scaled to $[0,1]$ (Section~\ref{sec:preprocessing}). Re-evaluating all models
on a group-aware, near-duplicate-free split with a common preprocessing
pipeline is ongoing work.

\textbf{Checkpoint used.} S2-A13 is a separate training run of the proposed
architecture, trained alongside the baselines on \texttt{phase2\_dataset14}
and before the positional-encoding ablation. The ablation checkpoint S2-A4,
trained only on the 655-image ablation corpus, scores mIoU 0.549 on this
benchmark: having never seen non-water scenes, it predicts algae on many of
the 116 algae-free test images. Retraining with the additional 476 images
(S2-A4$^{+}$, Section~\ref{sec:datasets}) reduces the mean false-positive area
on those images from 22.6\% to 4.9\% (SMP-UNet ResNet34: 5.0\%), but overall
mIoU rises only to 0.611, because 98 high-resolution drone images, which hold
95\% of the benchmark's pixels, are still over-segmented
(Section~\ref{sec:limitations}).

\textbf{EoMT-S.} EoMT-S (ViT-S/14 with DINOv2~\cite{oquab2024dinov2} pretraining) was fine-tuned
from its COCO-panoptic checkpoint with a much smaller budget (a 20-epoch run
followed by an 8-epoch classification-head fine-tune) on different hardware,
and its training pipeline requires every image to contain a target, so
purely-background images are excluded from its training and evaluation. It
has the lowest mIoU (0.7341), driven by weak background IoU (0.5726), while
its algae IoU (0.8955) is second only to the proposed model. Running the same
8-epoch fine-tune directly from the COCO-panoptic checkpoint gives lower
scores (mIoU 0.5857, background IoU 0.3146), suggesting that its main
limitation here is the short training schedule rather than the architecture.

\begin{table*}[htbp]
\caption{Comparison with baseline models on the 256-image test set, ranked by mIoU.}
\label{tab:model_comparison}
\centering
\resizebox{\textwidth}{!}{%
\begin{tabular}{lccccccc}
\toprule
\textbf{Model} & \textbf{mIoU} & \textbf{IoU Algae} & \textbf{IoU Bkg.} &
\textbf{Prec. Algae} & \textbf{Rec. Algae} & \textbf{Latency (ms)} & \textbf{Rank} \\
\midrule
SMP-UNet++ EfficientNet-B4       & \textbf{0.9239} & 0.8933 & \textbf{0.9546} & 0.9233 & \textbf{0.9649} & 26.90 & 1 \\
SMP-UNet ResNet34                & 0.9216 & 0.8896 & 0.9537 & 0.9313 & 0.9521 & 10.39 & 2 \\
DeepLabV3 ResNet101              & 0.9204 & 0.8873 & 0.9535 & 0.9415 & 0.9391 & 49.40 & 3 \\
Swin-UNet (Ours), S2-A13         & 0.9192 & \textbf{0.9021} & 0.9362 & \textbf{0.9499} & 0.9472 & 50.27 & 4 \\
SMP-FPN ResNet50                 & 0.9186 & 0.8848 & 0.9524 & 0.9396 & 0.9381 & 10.39 & 5 \\
SMP-DeepLabV3+ ResNet101         & 0.9166 & 0.8825 & 0.9507 & 0.9289 & 0.9464 & 15.95 & 6 \\
DeepLabV3 ResNet50               & 0.9164 & 0.8823 & 0.9505 & 0.9268 & 0.9484 & 32.89 & 7 \\
SMP-MAnet ResNet50               & 0.9093 & 0.8726 & 0.9461 & 0.9200 & 0.9442 & 32.69 & 8 \\
DeepLabV3 MobileNetV3            & 0.9027 & 0.8633 & 0.9420 & 0.9150 & 0.9385 & \textbf{6.48} & 9 \\
EoMT-S (fine-tuned)$^\dagger$     & 0.7341 & 0.8955 & 0.5726 & 0.9402 & 0.9496 & 20.38$^\ddagger$ & 10 \\
\bottomrule
\end{tabular}%
}
\vspace{2pt}
\begin{minipage}{\textwidth}\footnotesize
\textit{Ours} is checkpoint S2-A13 (see text).
$^\dagger$Smaller training budget; excludes purely-background images.
$^\ddagger$Measured on different hardware; not directly comparable.
\end{minipage}
\end{table*}

\begin{figure}[htbp]
\centerline{\includegraphics[width=\columnwidth]{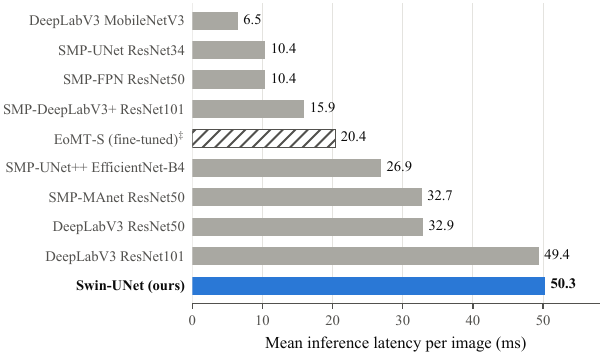}}
\caption{Mean inference latency per image on a single GPU (lower is faster).
$^{\ddagger}$Different hardware; not directly comparable.}
\label{fig:model_comparison_latency}
\end{figure}

Fig.~\ref{fig:model_comparison_latency} shows mean inference latency per
image on a single GPU. DeepLabV3 MobileNetV3 is the fastest model (6.48~ms);
the proposed model is the slowest of the directly comparable models
(50.27~ms), marginally behind DeepLabV3 ResNet101 (49.40~ms). Part of this
gap may reflect ONNX Runtime overhead, since the proposed model was timed as
an ONNX export and the baselines in native PyTorch. Its advantage in this
comparison is therefore in algae-class accuracy rather than speed.

%% ============================================================
\subsection{Controlled Ablation Study}
\label{sec:ablation}
%% ============================================================

\subsubsection{Design}

To isolate the contribution of each pipeline component, a controlled ablation
study was conducted across ten experiments covering three aspects of the
pipeline: Stage~1 water body pretraining quality (S1-A1 and S1-A2), Stage~2
transfer learning curriculum combined with dataset size effects
(S2-A1 through S2-A4$'$), and the two architectural enhancements introduced in
Section~\ref{sec:architecture} (S2-A5, S2-A8), each trained from the same
S1-A2 (RIWA+LaRS) checkpoint on the full expanded dataset. All experiments
use a single, identical random seed, backbone architecture, optimizer configuration,
and training duration. Each experiment changes exactly one variable from a
designated reference configuration. S2-A1$'$ and S2-A4$'$ are additional
baselines introduced to disentangle the effects of pretraining initialization
from annotated dataset size. Table~\ref{tab:ablation_design} summarizes the
experimental design.

\begin{table}[htbp]
\caption{Ablation study design. Primed IDs are added baselines that separate
dataset size from pretraining.}
\label{tab:ablation_design}
\begin{center}
\begin{tabular}{p{0.85cm}p{3.0cm}p{3.45cm}}
\hline
\textbf{ID} & \textbf{Experiment} & \textbf{Variable under test} \\
\hline
\multicolumn{3}{l}{\textit{Stage~1 --- Water body pretraining}} \\
S1-A1 & RIWA only             & RIWA as sole pretraining source \\
S1-A2 & RIWA + LaRS           & Sequential curriculum (reference) \\
\hline
\multicolumn{3}{l}{\textit{Stage~2 --- Transfer curriculum \& dataset size}} \\
S2-A1  & ImageNet only        & No pretraining, small data \\
S2-A1$'$ & Random init, small & No init, no expansion (lower bound) \\
S2-A2  & RIWA checkpoint      & RIWA pretraining, small data \\
S2-A3  & RIWA+LaRS checkpoint & Full curriculum, small data \\
S2-A4  & RIWA+LaRS + expanded & Full curriculum + expansion (ref.) \\
S2-A4$'$ & ImageNet + expanded & Expansion only, no pretraining \\
\hline
\multicolumn{3}{l}{\textit{Stage~2 --- Architectural enhancements (expanded dataset)}} \\
S2-A5  & No decoder attention & SCSE attention removed (\texttt{Identity}) \\
S2-A8  & No positional encoding & Learnable SPE module removed \\
\hline
\end{tabular}
\end{center}
\end{table}

\subsubsection{Stage~1 Ablation Results}

Table~\ref{tab:ablation_stage1} reports water body segmentation performance for
each Stage~1 variant. The near-ceiling mIoU achieved by S1-A2 (0.9899) confirms
that the RIWA+LaRS curriculum successfully trains a general-purpose aquatic
feature extractor, establishing the quality of representations available for
transfer into Stage~2.

\begin{table}[htbp]
\caption{Stage~1 ablation results at best validation checkpoint.
$^\dagger$S1-A1 on RIWA validation split (167 images);
S1-A2 on LaRS validation split (198 images).}
\label{tab:ablation_stage1}
\begin{center}
\begin{tabular}{lcccc}
\hline
\textbf{ID} & \textbf{mIoU} & \textbf{IoU$_\text{water}$} &
\textbf{IoU$_\text{BG}$} & \textbf{Pix.~Acc} \\
\hline
S1-A1: RIWA only$^\dagger$ & 0.9295 & 0.8975 & 0.9562 & 0.9697 \\
S1-A2: RIWA+LaRS (ref.)    & 0.9899 & 0.9884 & 0.9912 & 0.9950 \\
\hline
\end{tabular}
\end{center}
\end{table}

\subsubsection{Stage~2 Ablation Results}
\label{sec:stage2_ablation}

Table~\ref{tab:ablation_stage2} reports algae segmentation performance for all
ten Stage~2 experiments.

\begin{table*}[htbp]
\caption{Stage~2 ablation results (41-image validation split; bold: best per
column). $^\dagger$Trained on the 145-image corpus; all others on the
655-image corpus.}
\label{tab:ablation_stage2}
\begin{center}
\begin{tabular}{llccccc}
\hline
\textbf{ID} & \textbf{Experiment} &
\textbf{mIoU} & \textbf{IoU$_\text{algae}$} & \textbf{IoU$_\text{BG}$} &
\textbf{Pix.~Acc} & \textbf{Val Loss} \\
\hline
\multicolumn{7}{l}{\textit{Small dataset (145 images) --- pretraining comparison}} \\
S2-A1$^\dagger$   & ImageNet only          & 0.7450 & 0.7947 & 0.6953 & 0.8602 & 0.6019 \\
S2-A1$'^\dagger$  & Random init            & 0.7434 & 0.7942 & 0.6926 & 0.8594 & 0.6014 \\
S2-A2$^\dagger$   & RIWA checkpoint        & 0.7178 & 0.7684 & 0.6632 & 0.8420 & 0.5757 \\
S2-A3$^\dagger$   & RIWA+LaRS checkpoint   & 0.6609 & 0.7189 & 0.6010 & 0.8049 & 0.7446 \\
\hline
\multicolumn{7}{l}{\textit{Expanded dataset (655 images) --- dataset size and pretraining}} \\
S2-A4$'$          & ImageNet only          & 0.8508 & 0.8628 & 0.8282 & 0.9203 & 0.4725 \\
S2-A4             & RIWA+LaRS checkpoint (ref.) & 0.8614 & 0.8771 & 0.8456 & 0.9265 & 0.4473 \\
\hline
\multicolumn{7}{l}{\textit{Expanded dataset --- architectural ablations (from S2-A4 reference)}} \\
S2-A5             & No decoder attention (SCSE) & 0.8588 & 0.8730 & 0.8446 & 0.9249 & 0.4697 \\
S2-A8             & No positional encoding & \textbf{0.8676} & \textbf{0.8826} & \textbf{0.8525} & \textbf{0.9301} & \textbf{0.4307} \\
\hline
\end{tabular}
\end{center}
\end{table*}

\subsubsection{Discussion of Ablation Findings}
\label{sec:ablation_discussion}

All Stage~2 results come from a single training seed, evaluated at the best
epoch on the same 41-image validation split used for model selection. The
\mbox{$\sim$11}-point effect of dataset expansion is large relative to this
uncertainty, but differences of about one mIoU point or less (pretraining on
the expanded corpus, SCSE, positional encoding) are within typical run-to-run
variation and should be read as indicative only.

\textbf{Dataset size is the dominant performance driver.}
The most striking result in Table~\ref{tab:ablation_stage2} is the contrast
between the small-data group (S2-A1 to S2-A3, 145 images) and the expanded-data
group (S2-A4$'$ and S2-A4, 655 images). Expanding the annotated corpus from 145
to 655 images --- a 4.5$\times$ increase --- yields an mIoU gain of approximately
11 percentage points ($0.7450 \to 0.8614$), regardless of initialization. This
is the largest single performance improvement observed across all ablation
conditions by a wide margin.

\textbf{Water body pretraining is data-dependent: it hurts on small data but
shows directional advantage as annotation grows.}
Contrary to the central hypothesis of the three-stage pipeline, the RIWA+LaRS
water pretraining checkpoint does not consistently outperform ImageNet
initialization on the small dataset. S2-A2 (RIWA checkpoint, mIoU 0.7178) and
S2-A3 (RIWA+LaRS checkpoint, mIoU 0.6609) are both \textit{worse} than S2-A1
(ImageNet only, mIoU 0.7450). S2-A3 converged at only epoch 8 with elevated
validation loss (0.7446), indicating optimization instability rather than
productive adaptation when the LaRS-pretrained encoder encounters only 145 algae
training images. On the expanded dataset, however, the picture shifts: the
margin between S2-A4 (mIoU 0.8614) and S2-A4$'$ (ImageNet only, mIoU 0.8508)
is 0.0106 mIoU (1.06 percentage points) --- modest at this dataset scale, yet directionally consistent
with the hypothesis that aquatic domain pretraining becomes increasingly
advantageous as target-domain annotation grows. The implication is that water
body pretraining is not ineffective in principle, but rather requires a
sufficient volume of target-domain labeled data before the encoded aquatic
representations can be productively adapted to the fine-grained intra-water
discrimination the algae task demands.

\textbf{The role of the random initialization baseline.}
S2-A1$'$ (random initialization, 145 images, mIoU 0.7434) performs essentially
identically to S2-A1 (ImageNet initialization, 145 images, mIoU 0.7450),
suggesting that, under the current small-data configuration, the benefit of
pretrained representations is limited relative to optimization dynamics and
dataset constraints. The gradient signal from 145 annotated images appears
insufficient to either exploit ImageNet features productively or overcome
potential negative transfer from water pretraining. This result establishes
a practical lower bound for the small-data regime.

\textbf{Why pretraining does not help: the semantic gap hypothesis.}
This finding is counterintuitive given the near-perfect Stage~1 performance
(mIoU 0.9899) but can be explained by the fundamental nature of the task
transition. Binary water/non-water segmentation teaches the encoder to
discriminate the \textit{presence} of water against land, sky, and vegetation.
Algae segmentation, by contrast, requires discriminating \textit{within} a water
surface --- distinguishing bloom-covered pixels from open water pixels that are
spectrally and texturally similar and lie in the same spatial context. These are
fundamentally different discriminative tasks: the features that make water
detectable against land are not the features that make algae detectable against
open water. Encoder representations optimized for water detection may therefore
interfere with, rather than support, the fine-grained intra-water discrimination
required for algae segmentation --- a form of negative transfer well documented in
transfer learning literature~\cite{wang2019negative}.

\textbf{Practical implication: annotation first, pretraining compounds.}
The $\sim$11 percentage point mIoU gain from dataset expansion substantially
exceeds the pretraining contribution at the current annotation scale. For
practitioners with limited resources, the highest-return immediate investment is
acquiring and annotating additional target-domain images. However, the
directional advantage of RIWA+LaRS pretraining on the expanded dataset suggests
that pretraining and annotation are complementary: as the corpus grows, the
encoded aquatic representations are expected to provide increasing value, making
the pretraining pipeline a worthwhile long-term investment.

\textbf{Architectural enhancements: SCSE attention helps marginally, the
positional encoding module does not.}
Table~\ref{tab:ablation_stage2} isolates the two architectural additions
described in Section~\ref{sec:architecture}. Removing decoder SCSE
attention (S2-A5) reduces mIoU from 0.8614 to 0.8588 (0.26 percentage points),
consistent with SCSE providing a small benefit from recalibrating
channel and spatial responses in the decoder, as hypothesized. Removing the
learnable spatial positional encoding module (S2-A8), however, \textit{improves}
mIoU from 0.8614 to 0.8676 (0.62 percentage points) --- the best result of any
Stage~2 configuration tested, on every metric. This is the opposite of the
architectural motivation given in Section~\ref{sec:architecture}: on the
current dataset scale, the positional encoding module is not merely
unhelpful but mildly counterproductive, plausibly because a global
coordinate prior adds little value for a texture- and color-driven
discrimination task like algae-vs-water, while its added parameters and
gradient path introduce a small amount of extra optimization difficulty at
this annotation budget. The module is therefore omitted from the final configuration
(Section~\ref{sec:architecture}).

%% ============================================================
\subsection{Limitations}
\label{sec:limitations}
%% ============================================================

\textit{Benchmark overlap.} The 256-image comparison benchmark shares many
near-duplicate images with the comparison training pool
(Section~\ref{sec:model_comparison}), so the scores in
Table~\ref{tab:model_comparison} are likely optimistic.

\textit{Two checkpoints.} The ablation (S2-A4 and variants) and the model
comparison (S2-A13) use different training runs and different data pools, so
no single checkpoint is evaluated on both benchmarks.

\textit{Preprocessing.} The proposed model was trained on inputs scaled to
$[0,1]$, whereas the baselines used ImageNet mean/standard-deviation
normalization; the ImageNet-pretrained encoder may therefore be
disadvantaged.

\textit{Drone imagery.} High-resolution drone photographs are downsampled to
$512 \times 512$, which discards the fine texture that separates patchy
algae from clear water; the resulting over-segmentation of drone images
dominates pixel-weighted metrics on the comparison benchmark.

\textit{Statistical power.} Ablation results are single-seed, selected and
evaluated on a 41-image validation split.

%% ============================================================
\section{Conclusion}
\label{sec:conclusion}
%% ============================================================

This work presented a transfer-learning pipeline with attention-enhanced
decoding and model-assisted annotation for algal bloom segmentation in
freshwater imagery. In a controlled ablation, annotated dataset size was the
dominant factor: expanding the corpus from 145 to 655 images improved mIoU by
about 11 points. Water-body pretraining underperformed ImageNet
initialization on the small corpus and gave a small gain (1.06 points,
single seed) on the expanded corpus. SCSE attention gave a small gain and
the learnable positional encoding a small loss, so the latter was dropped.
For practitioners, annotation effort offers the clearest return at this
scale. Adding algae-free and non-water images to training reduced
false-positive area on non-water scenes from 22.6\% to 4.9\%. Against nine
baselines, a separately trained checkpoint of the proposed architecture
(S2-A13) achieved the highest algae IoU (0.9021) and fourth-highest mIoU
(0.9192), subject to the benchmark-overlap caveat above.

Future work will (1)~re-evaluate the proposed model and all baselines on a
group-aware, near-duplicate-free split with a common preprocessing pipeline,
reporting a single checkpoint across all benchmarks; (2)~repeat the
ablations over multiple seeds and report mean $\pm$ standard deviation;
(3)~address drone-image over-segmentation with tiled, higher-resolution
training and inference; (4)~use uncertainty-guided active learning to grow
the annotated corpus; and (5)~compress the model for on-board UAV deployment,
e.g.\ via quantization~\cite{gholami_survey_2021}.

\bibliographystyle{IEEEtran}
\bibliography{references/Computer_Vision,references/algae_intro_references,references/added_references}

\end{document}